\documentclass[letterpaper, 10 pt, conference]{ieeeconf}  

\IEEEoverridecommandlockouts                              

\usepackage{graphics} 
\usepackage{epsfig} 
\usepackage{mathptmx} 
\usepackage{times} 
\usepackage{amsmath} 
\usepackage{amssymb}  
\usepackage{bm}
\usepackage{multirow}
\usepackage{graphicx}
\usepackage[final]{changes}
\usepackage{optidef}
\usepackage{array}
\usepackage{colortbl}
\DeclareGraphicsExtensions{.pdf,png}
\title{\LARGE \bf
\added{Imitating and Finetuning Model Predictive Control for Robust and Symmetric Quadrupedal Locomotion}
}

\author{Donghoon Youm$^{*,1}$, Hyunyoung Jung$^{*,2}$, Hyeongjun Kim$^{1}$, Jemin Hwangbo$^{1}$, Hae-Won Park$^{1}$ and Sehoon Ha$^{2}$
\thanks{*These two authors contribute equally to work.}
\thanks{$^{1}$Korea Advanced Institute of Science and Technology, Yuseong-gu,
        Daejeon, 34141, Republic of Korea
        {\tt\small {ydh0725, kaist0914, jhwangbo, haewonpark}@kaist.ac.kr}}%
\thanks{$^{2}$Georgia Institute of Technology, Atlanta,
        GA, 30308, USA
        {\tt\small {hjung331, sehoonha}@gatech.edu}}%
\thanks{This paper has been accepted for publication in IEEE Robotics and Automation Letters (RA-L). 
The final version is available at https://doi.org/10.1109/LRA.2023.3320827
}
}
\begin{document}
\bstctlcite{IEEEexample:BSTcontrol}

\maketitle
\thispagestyle{empty}
\pagestyle{empty}



\newcommand{\cmt}[1]{}

\newcommand{\updated}[1]{\textcolor{blue}{{#1}}}
\newcommand{\hyunyoung}[1]{\textcolor{orange}{{Hyunyoung: #1}}}
\newcommand{\donghoon}[1]{\textcolor{cyan}{{Donghoon: #1}}} 
\newcommand{\sehoon}[1]{\textcolor{red}{{Sehoon: #1}}}

\newcommand{\newtext}[1]{#1}
\newcommand{\eqnref}[1]{Equation~(\ref{eq:#1})}
\newcommand{\figref}[1]{Figure~\ref{fig:#1}}
\newcommand{\tabref}[1]{Table~\ref{tab:#1}}
\newcommand{\secref}[1]{Section~\ref{sec:#1}}

\long\def\ignorethis#1{}

\newcommand{\etal}{{\em{et~al.}\ }}
\newcommand{\eg}{e.g.\ }
\newcommand{\ie}{i.e.\ }

\newcommand{\figtodo}[1]{\framebox[0.8\columnwidth]{\rule{0pt}{1in}#1}}



\newcommand{\pdd}[3]{\ensuremath{\frac{\partial^2{#1}}{\partial{#2}\,\partial{#3}}}}

\newcommand{\mat}[1]{\ensuremath{\mathbf{#1}}}
\newcommand{\set}[1]{\ensuremath{\mathcal{#1}}}

\newcommand{\vc}[1]{\ensuremath{\mathbf{#1}}}
\newcommand{\vEndEff}{\ensuremath{\vc{d}}}
\newcommand{\vRelMove}{\ensuremath{\vc{r}}}
\newcommand{\sSet}{\ensuremath{S}}

\newcommand{\vControl}{\ensuremath{\vc{u}}}
\newcommand{\vPoint}{\ensuremath{\vc{p}}}
\newcommand{\sSpringCoef}{{\ensuremath{k_{s}}}}
\newcommand{\sDamperCoef}{{\ensuremath{k_{d}}}}
\newcommand{\vHandle}{\ensuremath{\vc{h}}}
\newcommand{\vForce}{\ensuremath{\vc{f}}}

\newcommand{\mTransChain}{\ensuremath{\vc{W}}}
\newcommand{\mRotateTrans}{\ensuremath{\vc{R}}}
\newcommand{\sJoint}{\ensuremath{q}}
\newcommand{\vJoint}{\ensuremath{\vc{q}}}
\newcommand{\mJoint}{\ensuremath{\vc{Q}}}
\newcommand{\mMass}{\ensuremath{\vc{M}}}
\newcommand{\sMass}{\ensuremath{{m}}}
\newcommand{\vGravity}{\ensuremath{\vc{g}}}
\newcommand{\vConstr}{\ensuremath{\vc{C}}}
\newcommand{\sConstr}{\ensuremath{C}}
\newcommand{\vCOM}{\ensuremath{\vc{x}}}
\newcommand{\sGeneralForce}[1]{\ensuremath{Q_{#1}}}
\newcommand{\vStateVar}{\ensuremath{\vc{y}}}
\newcommand{\vControlVar}{\ensuremath{\vc{u}}}
\newcommand{\tr}[1]{\ensuremath{\mathrm{tr}\left(#1\right)}}

%
%

\renewcommand{\choose}[2]{\ensuremath{\left(\begin{array}{c} #1 \\ #2 \end{array} \right )}}

\newcommand{\gauss}[3]{\ensuremath{\mathcal{N}(#1 | #2 ; #3)}}

\newcommand{\pctab}{\hspace{0.2in}}
\newenvironment{pseudocode} {\begin{center} \begin{minipage}{\textwidth}
                             \normalsize \vspace{-2\baselineskip} \begin{tabbing}
                             \pctab \= \pctab \= \pctab \= \pctab \=
                             \pctab \= \pctab \= \pctab \= \pctab \= \\}
                            {\end{tabbing} \vspace{-2\baselineskip}
                             \end{minipage} \end{center}}
\newenvironment{items}      {\begin{list}{$\bullet$}
                              {\setlength{\partopsep}{\parskip}
                                \setlength{\parsep}{\parskip}
                                \setlength{\topsep}{0pt}
                                \setlength{\itemsep}{0pt}
                                \settowidth{\labelwidth}{$\bullet$}
                                \setlength{\labelsep}{1ex}
                                \setlength{\leftmargin}{\labelwidth}
                                \addtolength{\leftmargin}{\labelsep}
                                }
                              }
                            {\end{list}}
\newcommand{\newfun}[3]{\noindent\vspace{0pt}\fbox{\begin{minipage}{3.3truein}\vspace{#1}~ {#3}~\vspace{12pt}\end{minipage}}\vspace{#2}}

\newcommand{\key}{\textbf}
\newcommand{\fun}{\textsc}


\begin{abstract}

Control of legged robots is a challenging problem that has been investigated by different approaches, such as model-based control and learning algorithms. This work proposes a novel \added{Imitating and Finetuning Model Predictive Control (IFM)} framework to take the strengths of both approaches. Our framework first develops a conventional model predictive controller (MPC) using Differential Dynamic Programming and Raibert heuristic, which serves as an expert policy. Then we train a clone of the MPC using imitation learning to make the controller learnable. Finally, we leverage deep reinforcement learning with limited exploration for further finetuning the policy on more challenging terrains. By conducting comprehensive simulation and hardware experiments, we demonstrate that the proposed \added{IFM} framework can significantly improve the performance of the given MPC controller \replaced{on rough, slippery, and conveyor terrains}{, particularly on stairs and slippery terrains} that require careful coordination of footsteps. We also showcase that \added{IFM} can \added{efficiently} produce more symmetric, periodic, and energy-efficient gaits compared to Vanilla RL with a minimal burden of reward shaping.

\end{abstract}

\section{introduction}
Designing a robust locomotion controller has been a challenging problem for roboticists due to the under-actuated, discrete, and high-dimensional dynamics of legged robots. Roboticists have tackled legged locomotion with various approaches, from model-based control methods to learning-based techniques. Over the decades, traditional model-based control has demonstrated impressive agile and robust gaits on quadrupedal robots~\cite{hong2020real,ding2021representation,di2018dynamic,gehring2013control,bledt2018cheetah,katz2019mini,kim2019highly}. However, it often requires extensive effort to develop proper mathematical models, which becomes more challenging for complex and unconventional situations. On the other hand, learning-based approaches, such as deep reinforcement learning (deep RL), have the potential to automatically solve unintuitive locomotion problems on challenging terrains by leveraging a massive amount of data~\cite{hwangbo2019learning,ji2022concurrent,yang2020multi,yu2020learning,kumar2021rma,miki2022learning,margolisyang2022rapid,rudin2022learning}. However, deep RL may need to be followed by repetitive reward shaping to obtain high-quality motions. Otherwise, the learned policy may learn jerky and asymmetric gaits that cannot be transferred to real robots.

In recent years, researchers have investigated frameworks to combine model-based control and deep RL, taking the strengths of both approaches. One popular method is a hierarchical control architecture, where a learnable policy generates high-level commands, such as the target speed or footstep locations, for a low-level model predictive controller (MPC)~\cite{xie2023glide,yang2022fast,yu2021visual}. Also, there have been attempts to replace one control module with a neural network~\cite{da2021learning}. However, these approaches still require executing MPC at high frequency and can yield additional issues, such as significantly increased simulation costs and inference speed.

\begin{figure}
\centering
\includegraphics[trim={0 1cm 0 1cm},clip,width=0.9\columnwidth]{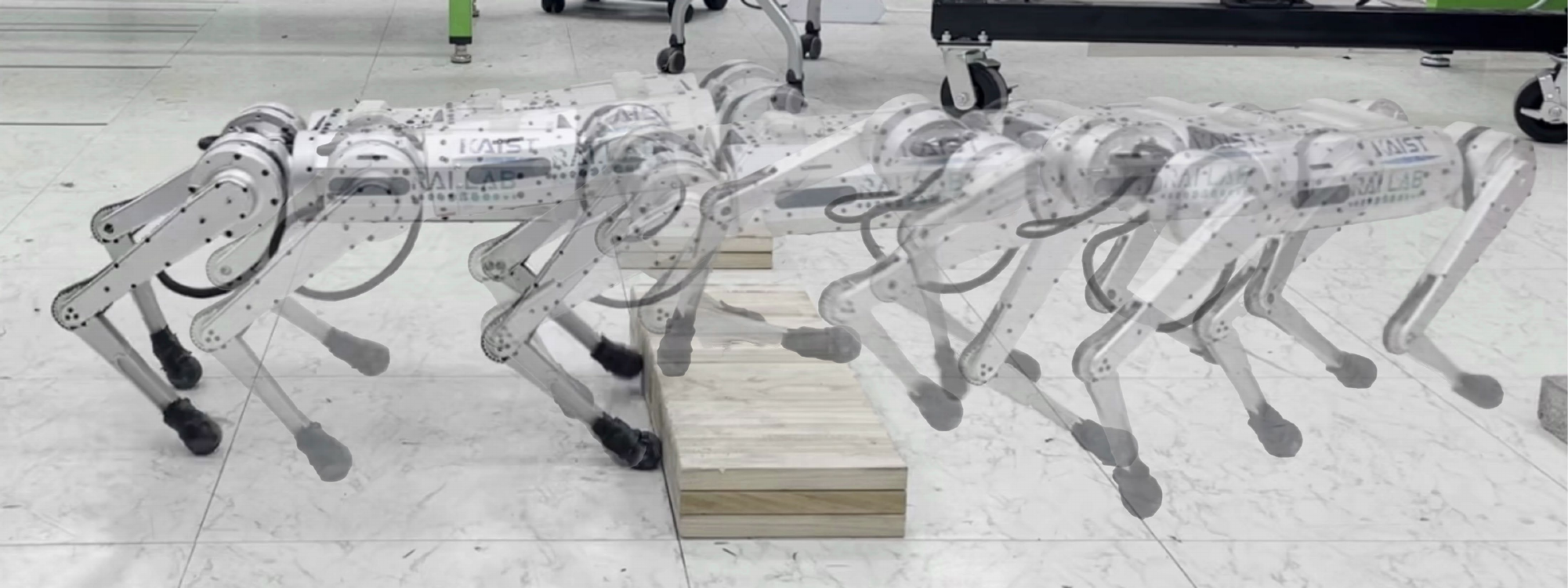}
\vspace{-0.5em}
\caption{Our \added{Imitating and Finetuning Model Predictive Control 
 (IFM)} framework can learn a robust, symmetric, periodic, and efficient policy that enables the Mini-cheetah robot to overcome a 7.5cm height obstacle.}
\label{fig:teaser}
\vspace{-0.5cm}
\end{figure}

We seek inspiration from the recent success of pre-training in various machine learning areas, including computer vision and natural language processing.
Particularly, we pre-train a policy by mimicking the expert's behaviors. One possible solution is to replicate the generated actions of the MPC using supervised learning by carefully designing a loss function~\cite{carius2020mpc,reske2021imitation}. On the other hand, motion imitation allows users to learn a policy that can track the given reference motion using deep RL, which shows natural motions both in simulation~\cite{peng2018deepmimic} and on hardware~\cite{RoboImitationPeng20,fuchioka2022opt}. 
Bogdanovic et al.~\cite{bogdanovic2021model} further proposed to improve the performance via task-based finetuning.
However, the performance of motion imitation is still bounded by the given single reference trajectory, which makes it difficult to obtain a general policy for various command tracking tasks.

This work proposes an \added{Imitating and Finetuning Model Predictive Control~(IFM)} framework, which is designed to take advantage of both model-based control and learning-based algorithms. We first design an MPC using Differential Dynamic Programming (DDP) and Raibert heuristics. We then train a neural network policy to imitate the expert controller using Dataset Aggregation (DAgger)~\cite{ross2011reduction}. Finally, we finetune the policy to overcome various challenging terrains using the on-policy deep RL algorithm, Proximal Policy Optimization~\cite{schulman2017proximal}. We further employ many sub-components, such as action space conversion, constrained exploration, terrain curriculum, and domain randomization, for seamless integration and better performance. 

We demonstrate the benefit of the proposed \added{IFM} framework by conducting both simulation and hardware experiments on the Mini Cheetah robot~\cite{katz2019mini}. We first show that our approach can train a new policy with improved robustness compared to the expert policy (MPC) on rough terrains, particularly when it is required to take unconventional footfall patterns, such as moving conveyor belts. It also has significantly reduced computation time compared to that of MPC. In addition, \added{IFM} finds symmetric, periodic, and energy-efficient gaits compared to Vanilla RL with a minimal burden of reward shaping, \added{which leads to better hardware performance.} It also converges much faster than Vanilla RL, which makes it more suitable for challenging locomotion problems. In our experiments, the \added{IFM} policies outperform the other methods on hardware regarding the success rate and tracking error. We summarize the contributions as follows. 
\begin{itemize}
\item We propose a novel \added{Imitating and Finetuning Model Predictive Control~(IFM)} framework to take the benefits of MPC and RL.
\item We demonstrate that \added{IFM} can significantly improve the performance of MPC and reduces the computation time.
\item We demonstrate that \added{IFM can improve the performance of Vanilla RL in terms of overall tracking errors and motion styles}, with less burden of reward engineering.
\end{itemize}

\section{related work}
\subsection{Model-based approaches}
A model-based approach\deleted{ for legged robot control} leverages a mathematical model to illustrate the dynamics of the given system. In the case of quadrupedal robot control, the robot is typically modeled as a single rigid body with small leg mass assumption~\cite{hong2020real,ding2021representation,di2018dynamic,gehring2013control,bledt2018cheetah} to make optimization feasible. Particularly, MPC~\cite{hong2020real,ding2021representation,di2018dynamic,bledt2018cheetah,katz2019mini,kim2019highly} has been widely used for several decades. It predicts the robot's future states and generates optimal control inputs based on a predefined cost function. MPC has the advantages of being explainable, robust, and capable of producing natural locomotion in legged robots. However, MPC is often formulated based on the predefined model, \replaced{which requires experts' prior knowledge to develop specific models to overcome rough and dynamic terrains.}{which makes it difficult to control legged robots on rough and dynamic terrains that require rapid adaptation of foothold selections.} Some recent research has focused on using terrain information and model-based optimization for foothold selection~\cite{fankhauser2018robust}. Still, many researchers adopt Raibert's heuristics~\cite{raibert1986legged}, which cannot guarantee the robot's performance in cases where the foot slips significantly during walking or gets stuck on obstacles. In this work, we utilize the aforementioned advantages of MPC while compensating for its weakness with a learning-based approach.

\subsection{Learning-based approaches}
The recent progress in learning-based control methods has demonstrated the effectiveness of legged robot control~\cite{hwangbo2019learning,ji2022concurrent,yang2020multi}, which has been proven to be effective in many scenarios. The performance of these learning-based approaches can be further boosted by introducing additional techniques, such as privileged learning~\cite{lee2020learning}, online adaptation~\cite{yu2020learning,kumar2021rma}, system identification~\cite{hwangbo2019learning}, belief propagation~\cite{miki2022learning}, curriculum learning~\cite{margolisyang2022rapid,rudin2022learning}, and many more. However, learning-based approaches often require careful reward shaping to obtain high-quality, symmetric, and periodic motions, which sometimes need up to ten reward terms, ranging from task terms to various regularization. The proposed framework, \added{IFM}, intends to supplement these works by pre-training a control policy with an MPC expert.\replaced{ Liang et al.~\cite{liang2018cirl} employed a similar approach. However, their use of human demonstrations precludes the application of DAgger~\cite{ross2011reduction} and there are no results from the real-world test.} {We want to note that our work can be combined with most of the existing learning-based frameworks due to its simplicity.}

\subsection{Combined approaches}
There exist a few works that aim to combine model-based and learning-based control approaches. One approach is hierarchically integrating the MPC and RL algorithms, where high-level decision-making is done by the learnable policy, and low-level motion control is done by the model-based control~\cite{xie2023glide,yang2022fast,da2021learning,yu2021visual}. Villarreal et al.~\cite{villarreal2020mpc} adopted this approach for combining a vision-based foothold selector and MPC. The other approach is to train the residual network using a learnable policy together with a model-based controller~\cite{carroll2023agile}. While demonstrating significant performance improvement, these formulations\deleted{ still} require expensive optimization for running MPC.\deleted{ from the backend, which makes this approach computationally expensive.} Carius et al.~\cite{carius2020mpc} and Reske et al.~\cite{reske2021imitation} utilize the Hamiltonian function as a loss function to\deleted{ effectively} imitate the existing MPC. Their methods can even slightly outperform the original controller by learning from multiple experts, but still, there is no explicit notion of finetuning.

\added{The approach of imitating MPC has been explored in other applications, such as autonomous driving~\cite{Pan-RSS-18} and drone~\cite{Kaufmann-RSS-20,Tagliabue2022demonstration}. Particularly, Nagami et al.~\cite{Nagami-RSS-21} adopted this imitation learning as a pre-training to solve problems with sparse rewards that are hard to be solved by Vanilla RL. Inspired by these works, our work aims to significantly improve the performance of MPC-based locomotion controllers while improving the quality of the RL-generated motions without reward engineering. }

\begin{figure}[t]
\centering
\vspace{0.25cm}
\includegraphics[width=0.8\columnwidth]{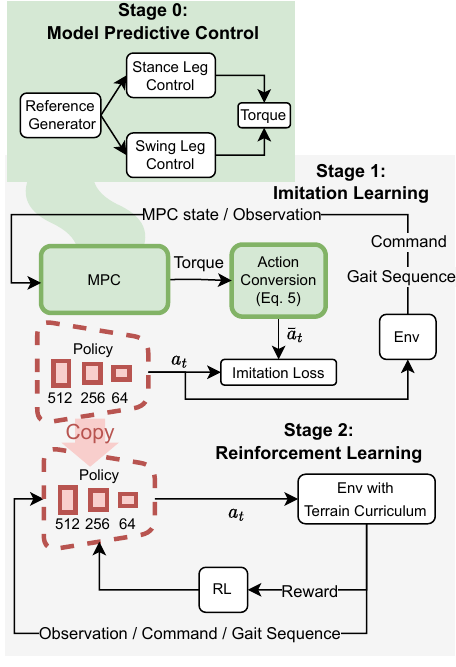}
\vspace{-0.3cm}
\caption{
Overview of the proposed \added{Imitating and Finetuning Model Predictive Control} (\added{IFM}). We first develop the expert controller $\pi^E$ using \replaced{MPC}{model-based algorithm}. Then we pre-train a neural network policy $\pi^I$ by cloning the expert's behaviors on the flat terrain. Finally, we finetune the policy on more challenging terrains\added{ using RL}\deleted{, while limiting the exploration to preserve the original style}.
}
\vspace{-0.5cm}
\label{fig:framework}
\end{figure}

\section{\added{Imitating and Finetuning MPC}}
\label{sec:method}

We present a novel \added{Imitating and Finetuning Model Predictive Control~(IFM)} framework for combining the strengths of model-based and learning-based control. We first develop an expert MPC using DDP and Raibert heuristics. Then we pre-train a policy by imitating the given expert MPC and finetune the cloned policy on various challenging tasks. The framework is summarized in Fig.~\ref{fig:framework}.

From the perspective of MPC, our framework provides an opportunity to improve the controller on challenging terrains that are hard to be described mathematically. From the perspective of deep RL, our framework finds periodic, symmetric, and physically plausible gaits without complex reward shaping in a sample efficient manner. 

\subsection{Stage 0: Model Predictive Control}
\label{sec:MPC}
Our framework, \added{IFM}, starts by developing an expert controller $\pi^E$. In theory, our framework is agnostic to the design process of the given expert policy, which can be obtained from various approaches, including heuristic methods, model-based algorithms, or other learning-based methods. In our implementation, we develop an expert policy, which uses DDP as in Tassa et al.~\cite{tassa2014control} with friction pyramid constraints for stance legs and Raibert heuristics~\cite{raibert1986legged} coupled with task space impedance control for swing legs.

\subsubsection{Controller Design}

We define the state and control input of the convex MPC as follows:
\begin{equation}
    \begin{split}
        \mathbf{x}(t) & \triangleq [\Theta_t, \mathbf{p}_t, \mathbf{\omega}_t, \mathbf{v}_t] \\
        \mathbf{u}(t) & \triangleq [ \mathbf{f}_1^T,\mathbf{f}_2^T,\mathbf{f}_3^T,\mathbf{f}_4^T],
    \end{split}
\end{equation}
where $\Theta_t$, $\mathbf{p}_t$, $\mathbf{\omega}_t$, and $\mathbf{v}_t$, are the Z-Y-X Euler angles, position, angular velocity, and linear velocity of the base at time $t$, respectively. Each $\mathbf{f}_i$ represents a ground reaction force of the $i^{th}$ foot and is multiplied by the transpose of foot Jacobian to apply joint torque.

Then we define our cost function of DDP as follows:
\begin{equation}
\label{eq:localcost}
\underset{\mathbf{x},\mathbf{u}}{\min}\sum_{i=0}^{T-1}\left\|\mathbf{x}_{i+1}- \mathbf{\overline{x}_{i+1}}\right\|_{\textbf{Q}} + \left\|\textbf{u}_i \right\|_\textbf{R},
\end{equation}    
where $\bar{\mathbf{x}}$ is the desired state and $\mathbf{Q}$ and $\mathbf{R}$ are weight matrices. The reference generator outputs a reference state trajectory by incorporating the current state and velocity command in the same manner as Di et al.~\cite{di2018dynamic}. The velocity command comprises the desired linear velocities in the x and y directions and a yaw velocity. 
In order to match the control frequency of the policy network, which will be described later, 100Hz, the control $dt$ of the MPC was set to 0.01s, and the horizon length T was set to 26. 

Note that the formulated MPC cannot be executed in real-time on hardware because it takes $12.382$ ms for optimization. However, we only run this MPC controller in simulation to generate pre-training data.

In addition, we made two changes to the original framework of Tassa et al.~\cite{tassa2014control} based on the work of Di et al.~\cite{di2018dynamic}. First, we adopt friction pyramid constraints instead of box constraints. We also implement the swing leg controller to compute joint torque for leg $i$ using task space impedance control as:
\begin{equation}
\label{eq:swing_leg_control}
    \tau_i = \mathbf{J}_i^T[K_p^t (\mathbf{p}^f_i - \bar{\mathbf{p}}^f_i) + K_d^t (\mathbf{v}^f_i - \bar{\mathbf{v}}^f_i)]
\end{equation}
where $\mathbf{J}_i$ is the $i$-th foot Jacobian, $\mathbf{p}^f_i, \mathbf{v}^f_i$ are the position and velocity of the $i$-th swing leg foot. We used task space PD gain $K_p^t=150$ and $K_d^t=5$. The upper bar denotes the reference foot state. The reference generator uses the Raibert heuristic~\cite{raibert1986legged} for the foothold and a cubic Bézier curve connecting the current foot to this foothold to create the reference foot trajectory. The gait sequence is a predefined swing time of $0.13$~s and a stance time of $0.13$~s without flying phases. 

\subsubsection{Action Conversion} 
Many MPC frameworks generate torques by converting the given desired contact forces using Jacobian transpose, including our expert controller.  
Conversely, it is more common for RL to select the target joint angles as actions because torque-based policies often exhibit significant vibration on hardware.
Therefore, we convert the torque command of the MPC into the target joint angles to make it compatible with imitation and RL policies.

For this purpose, we assume the joint space proportional-derivative~(PD) controller that generates torques as:
\begin{equation} \label{eq:pd}
    \tau = K_p^j (\mathbf{q} - \bar{\mathbf{q}}) - K_d^j \dot{\mathbf{q}},
\end{equation}
where $\mathbf{q}$, $\dot{\mathbf{q}}$, and $\bar{\mathbf{q}}$ are the current joint angles, the current joint velocities, and the target joint angles, respectively. We used joint space PD gains $K_p^j=17$ and $K_d^j=0.4$ in our whole experiments. We rearrange this equation to compute the target angle $\bar{\mathbf{q}}$ from the given torque $\tau$:
\begin{equation} \label{eq:inverse_pd}
    \bar{\mathbf{q}} = \frac{1}{K_p^j}(K_p^j \mathbf{q} - K_d^j \dot{\mathbf{q}} - \tau).
\end{equation}

\subsection{Stage 1: Imitation Learning} \label{sec:IL}
\subsubsection{Imitation learning with Data Aggregation}
Despite its robustness, the expert $\pi^E$ is intrinsically a fixed controller that prevents it from being trained for further improvement. In addition, it is computationally more expensive than a single feed-forward neural network because of complex numerical optimization. Therefore, we convert $\pi^E$ to the neural network policy $\pi^I$ whose weights can be optimized by minimizing the imitation loss:
\begin{equation} \label{eq:imitation}
    L_{imi} = \sum_{\mathbf{x} \in D}{ \left\| \pi^E(\mathbf{x}) - \pi^I(\mathbf{x}) \right\|^2},
\end{equation}
where $D$ is the collected dataset.
However, this naive behavior cloning approach often fails when the policy encounters previously unseen data during testing, 
which is known as the distribution mismatch problem. We can mitigate this issue by adopting the Dataset Aggregation~(DAgger) approach~\cite{ross2011reduction}, where we roll out ten trajectories from the policy $\pi^I$, label the correct action with the expert $\pi^E$ and add them to the dataset $D$ for every iteration. This approach results in a more stable policy by collecting in-distribution data and using it for training.
Note that this imitation stage is not computationally expensive compared to the RL finetuning: we only need to sample less than 400K data for DAgger training.

\subsubsection{Policy Architecture} 
The policy $\pi^I$ is implemented as a Multi-Layer Perceptron~(MLP) with three hidden layers consisting of 512, 256, and 64 neurons and leaky ReLU as the activation function. 
The input to the MLP consists of body orientation, body angular velocity, current joint angle and velocity, previous action, joint angle, and velocity history, predefined contact sequence and phase, and user command.
Although the predefined contact sequence is provided as part of inputs, the policy $\pi^I$ does not need to follow the fixed contact schedule explicitly. This implies that our policy is not strictly limited to fixed footfall patterns defined in Sec.~\ref{sec:MPC}.

The joint history is vital information for blind locomotion on rough terrain, which enables the robot to infer terrain properties. Although we only use flat terrain in the imitation learning stage, we include joint history as an input for later RL finetuning tasks.
We also do not include the body linear velocity as an observation since it can be implicitly inferred from previous outputs, joint history, and current joint information. In our experience, explicitly providing the body linear velocity as an input can potentially degrade the performance of the policy on hardware because the robot needs to rely on a noisy state estimator.

The policy's output is defined as the joint angle difference from its initial position, scaled by a factor $\sigma$, such that $\bar{\mathbf{q}} = \mathbf{q}_{init} + \sigma\mathbf{a}$, where $\mathbf{q}_{init}$ is the predefined initial pose, $\mathbf{a}$ is the policy's output, and $\sigma$ is the scaling factor.
\subsubsection{Training Details} 
We optimize the policy parameters with respect to the given loss function (Eq.~\ref{eq:imitation}) using adaptive momentum estimation~(Adam)~\cite{kingma2014adam}.
We terminate the sampling if the non-feet body part touches the ground or the robot flips over 70$^{\circ}$ to avoid including samples where the MPC has failed to converge. The batch size of data increases linearly by aggregating 4000 data at every iteration (10 environments $\times$ 400 transitions), and each batch is trained for 10 epochs.\\

\subsection{Stage 2: Reinforcement Learning} \label{sec:RL}
So far, we have obtained the initial expert controller $\pi^E$ and the imitation policy $\pi^I$. However, neither is optimal due to the model assumptions of the original expert. In this section, we aim to further boost the performance of the policy by finetuning it on challenging terrains using RL. Compared to the MPC, the RL policy $\pi^R$ can flexibly adapt the robot's gait to various terrains without explicit modeling. Compared to IL, RL explicitly optimizes the performance of the policy on downstream tasks. 
\subsubsection{Curriculum over Downstream Tasks}
\begin{figure}
\centering
\setlength{\tabcolsep}{0pt}
\renewcommand{\arraystretch}{0.}
\resizebox{0.9\columnwidth}{!}{%
\setlength{\arrayrulewidth}{1pt}
\begin{tabular}{cc}
\includegraphics[trim=0 0 0 3.5cm,clip,scale=1]{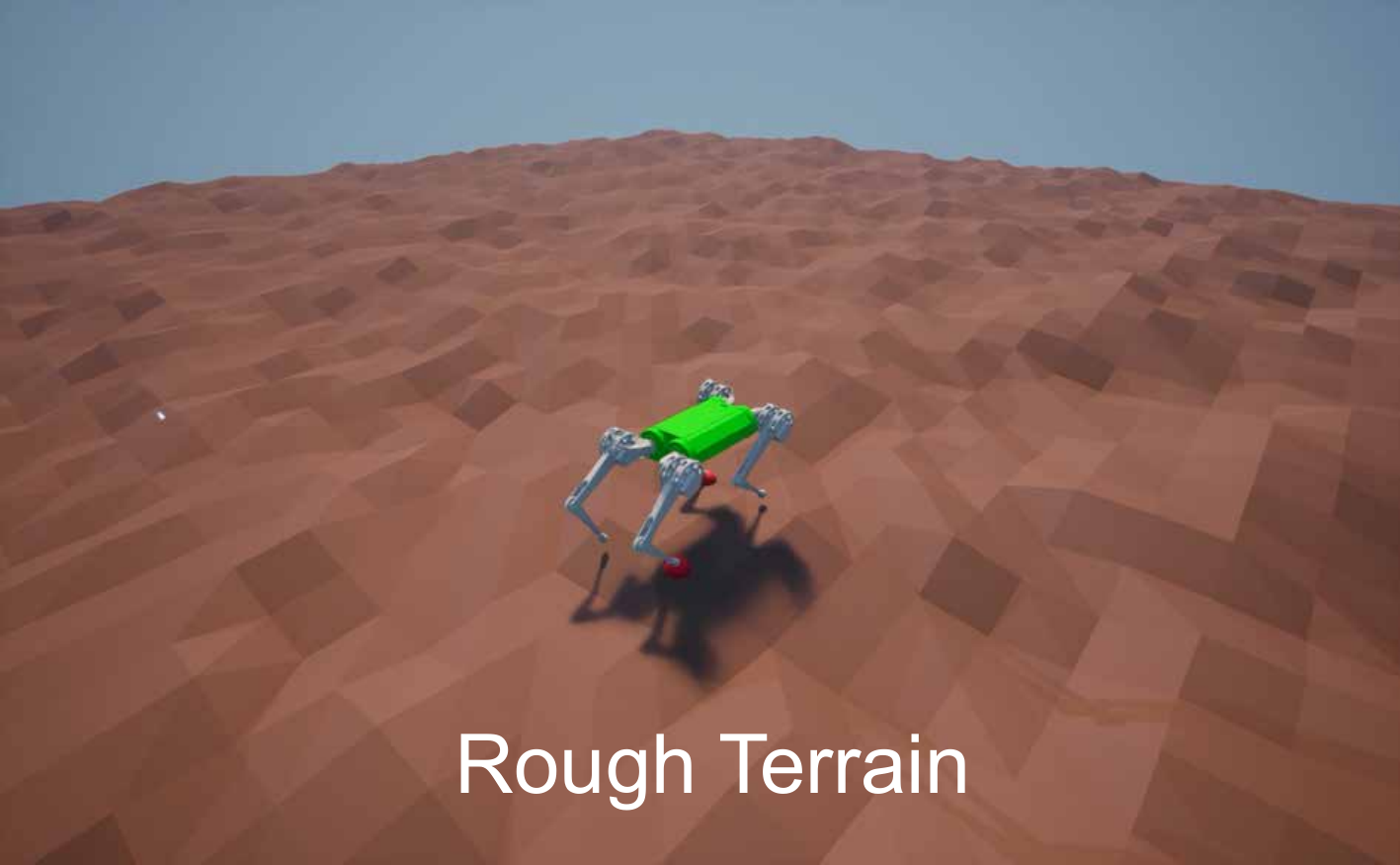} & \includegraphics[trim=0 0 0 3.5cm,clip,scale=1]{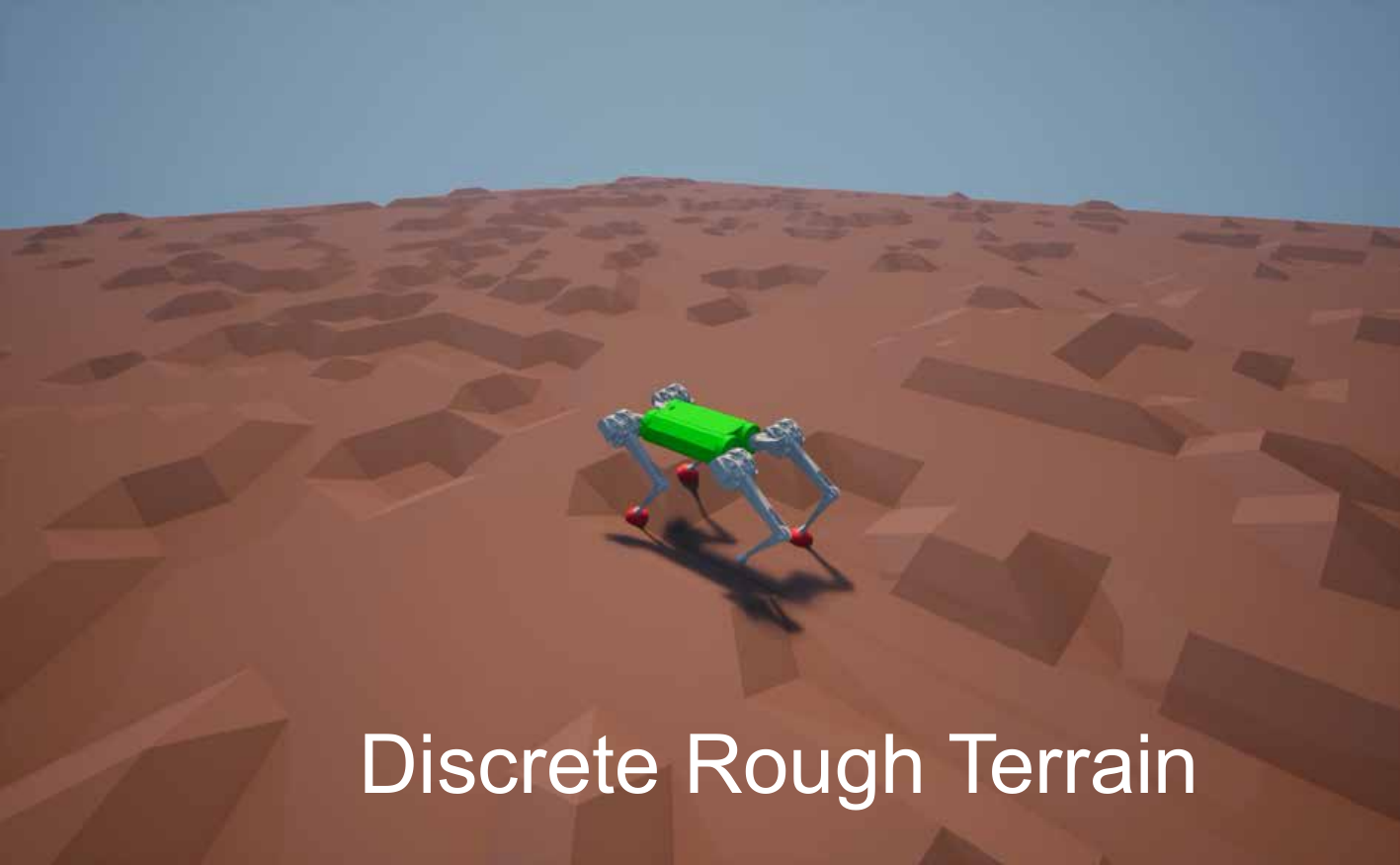} \\ 
\includegraphics[trim=0 0 0 3.5cm,clip,scale=1]{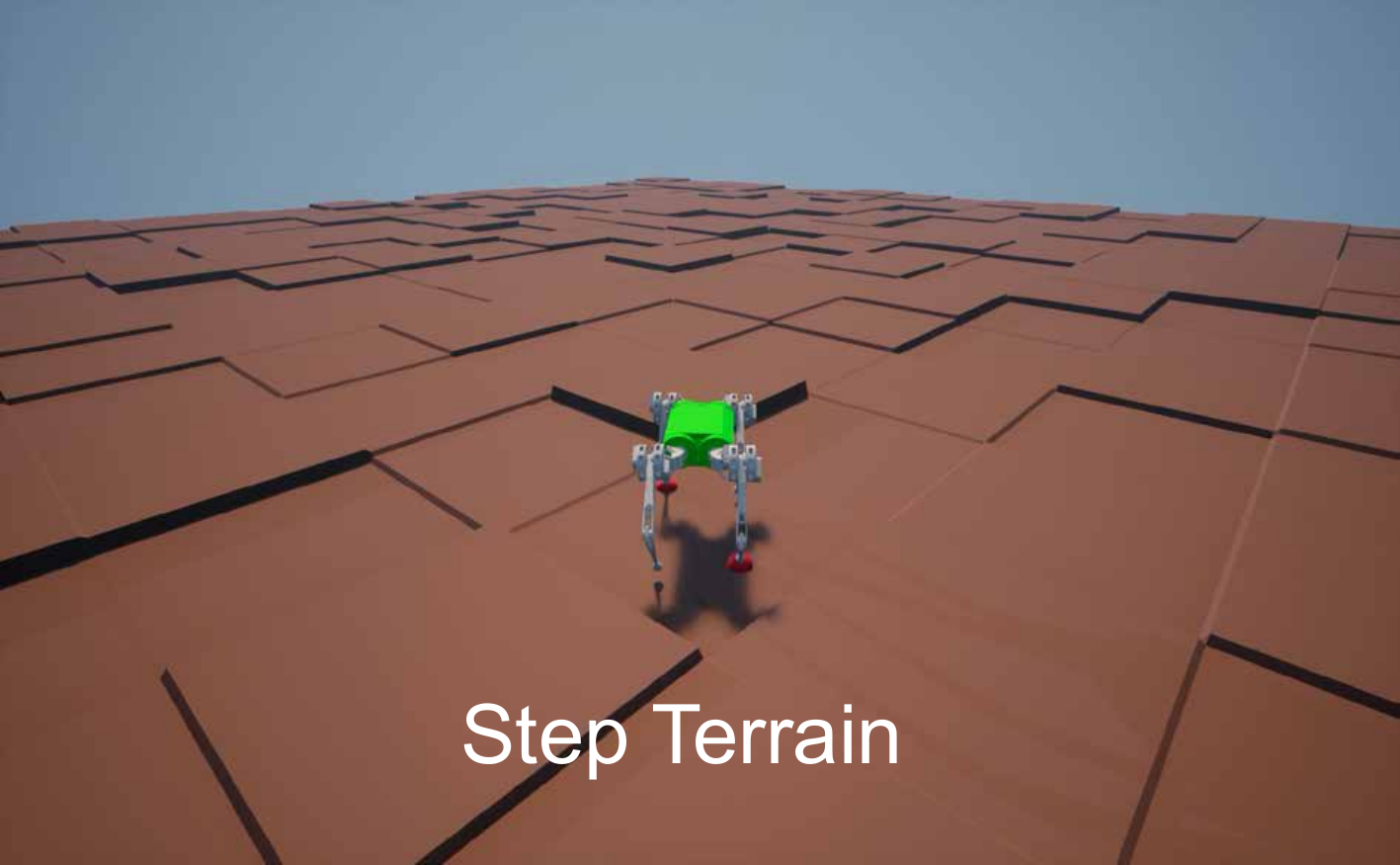} & \includegraphics[trim=0 0 0 3.5cm,clip,scale=1]{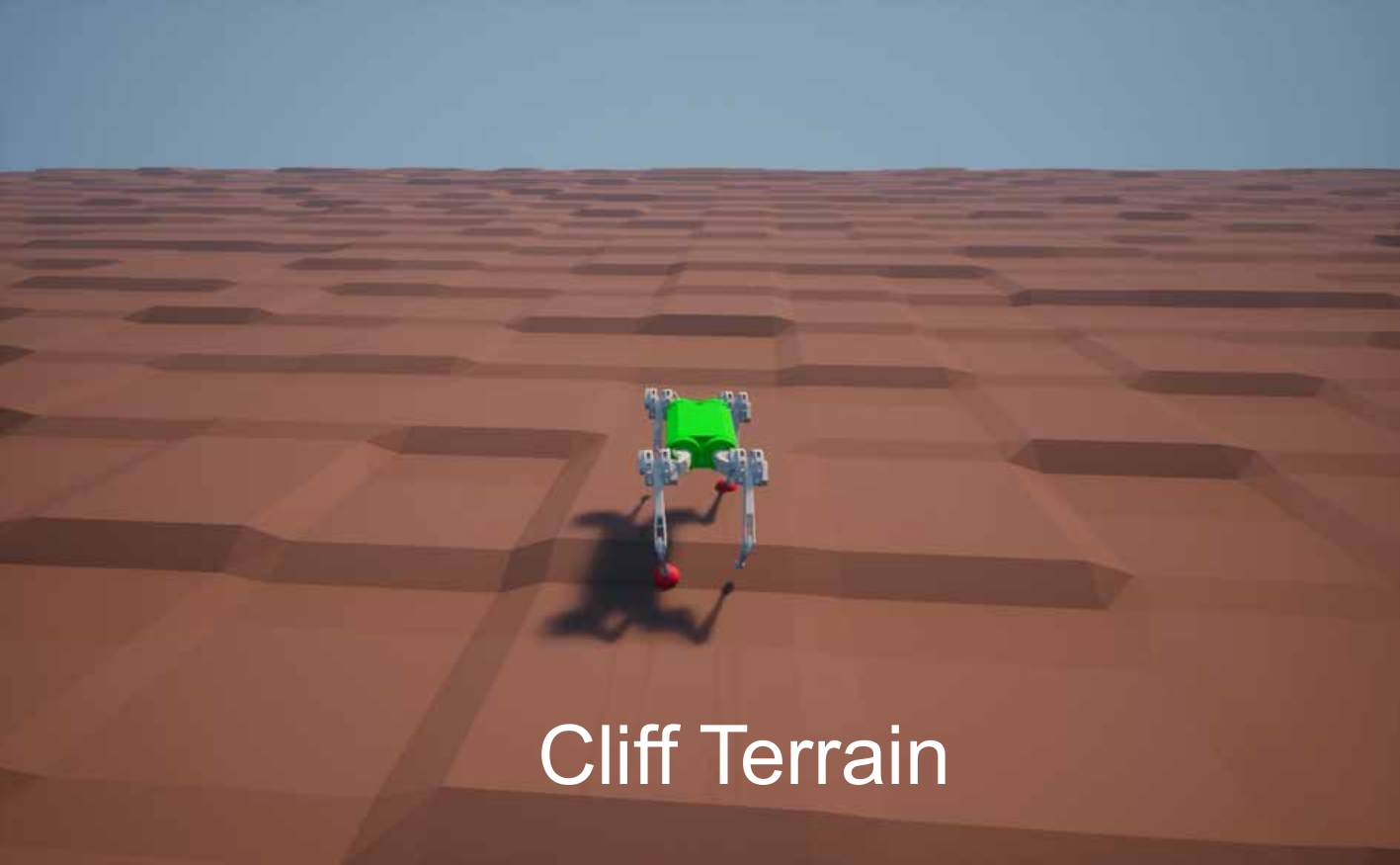}
\end{tabular}%
}
\vspace{-0.0cm}
\caption{Illustration of the challenging terrains for RL finetuning.}
\vspace{-0.5cm}
\label{fig:terrains}
\end{figure}
We select several challenging terrains as downstream tasks, including rough terrain, discrete rough terrain, step terrain, and cliff terrain, as shown in Fig.~\ref{fig:terrains}. These terrains are generated by varying the \emph{terrain factor}, which is the scaled maximum height of the terrain. \added{For example, we used Perlin noise with 0.5 fractal octaves, 1 to 5 fractal lacunarity and 0.45 fractal gain while varying the z-scale using this \emph{terrain factor} to generate rough terrain.} 
\replaced{We also introduce curriculum learning to avoid the loss of the style on harsh terrains.}{finetuning a pre-trained policy on harsh terrain at the early learning stage may cause the policy to lose its pre-trained motion. We use a curriculum approach to prevent this, gradually increasing the terrain difficulty.}
Specifically, we start with a terrain factor of 0.01 and increase it exponentially by a factor of 0.96 every 5 iterations.  We also sample the flat terrain with about a 5\% probability to prevent the robot from losing its pre-trained motion.

\subsubsection{Reward Formulation}
\label{sec:reward_formulation}

One challenge in applying deep RL to high-dimensional control problems is that the agent often produces {unintended} and noisy motions that cannot be transferred to a real robot, depending on the formulation.
Therefore, researchers often have approached this problem with manual reward engineering, which uses up to ten reward terms to obtain the desired style of locomotion~\cite{margolisyang2022rapid,rudin2022learning}.
On the other hand, we unburden this process by leveraging imitation learning which reduces the search space of policy to an action space that is feasible for the real robot. 

We finetune the policy using deep RL \added{in two different settings.} \added{In a simple reward (SR) setting, the following minimally constructed reward terms are used}\deleted{with the minimally constructed reward with three terms}: (1) tracking the given velocity command $[\bar{v}_{x},\bar{v}_y,\bar{\omega}_z]$, (2) minimizing the unnecessary movements in the other directions ($v_{z}, \omega_x, \omega_y$), and (3) regularizing the torque usage. \added{In a contact reward (CR) setting, we introduce an additional contact biasing reward on top of the existing terms to encourage the policy to follow a predefined contact sequence.}

For each episode, we also randomly sample the initial state, and the target command from $U(-1, 1)$.

\subsubsection{Domain Randomization}
The model trained on simulation might not work if we deploy it directly to the real world. Therefore, We perform domain randomization for successful zero-shot sim-to-real transfer.
We adopt a similar randomization technique as Ji et al.~\cite{ji2022concurrent}. To be more specific, we randomize the observation noise, motor friction, PD controller gains, ground friction coefficient, and the delay in motor command transmission time.
For the range of ground friction coefficient, we uniformly sample from $[0.3,1.0]$, which is a wider range than Ji et al.
Since there is about $10$~ms latency between the actual controller and the robot, we model this latency by setting the PD target after uniformly sampled time between $[0,10]$~ms.

\subsubsection{Training Details}
We use Proximal Policy Optimization~(PPO)~\cite{schulman2017proximal} coupled with  asymmetric actor-critic algorithm~\cite{pinto2017asymmetric} for training.
We use the same actor network as in Sec.~\ref{sec:IL} but modify the policy to have a Gaussian distribution to effectively explore possible solutions during training. To be more specific, the MLP structure outputs the mean of the policy, and the learnable parameters act as the standard deviation of the Gaussian distribution.
For the critic network, we use the same MLP structure as in the actor network, which has three hidden layers consisting of 512, 256, and 64 neurons and leaky ReLU as the activation function. 
In addition, unlike actor observation, the noise was not included in the value observation, but the linear velocity and body height were included, and joint history was excluded.

\subsubsection{Constrained Exploration}
In \added{IFM}, it is important for the policy to maintain its pre-trained motion during finetuning. Therefore, we set the initial standard deviation, learning rate, and PPO clip parameters to small values to allow the policy to explore the environment while retaining its meaningful pre-trained motion. Specifically, we set the initial standard deviation to $1.0$\added{~(or $2.0$ in CR)}, the policy network learning rate to $10^{-5}$, and the PPO clip parameter to $0.05$\added{~(or $0.1$ in CR)}. We use a learning rate of $10^{-3}$ for the critic network\deleted{to accelerate the learning process}. 

\section{Experimental Results}
We design experiments to investigate the following research questions: (1) can \added{IFM} improve the robustness of model-based controller and reduce the computation time? 
\replaced{(2) Can IFM improve the performance of Vanilla RL in terms of tracking errors and motion quality?}{(2) can \added{IFM} learn more symmetric, periodic, and energy-efficient gaits than Vanilla RL? }
(3) Is \added{IFM} more sample efficient than Vanilla RL? \\ 
\vspace{-0.5cm}
\subsection{Experimental Details}
\subsubsection{Baselines}
We select five controllers for our baselines which are given as follows:
\begin{itemize}
    \item \textbf{MPC:} We use two different MPCs for simulation and hardware experiments. In the simulation, MPC is the model-based control policy described in Sec.~\ref{sec:MPC}. In hardware experiments, the controller is convex MPC with the Whole Body Impulse Control (WBIC)~\cite{kim2019highly} that is tuned for the hardware.
    \item \textbf{DAgger:} DAgger is the policy trained up to the imitation stage in Sec.~\ref{sec:IL}.
    \item \textbf{Vanilla RL\added{(SR)}:} The Vanilla RL\added{(SR)} baseline is trained using the same reward function as the \replaced{simple reward}{Vanilla} setting in Sec.~\ref{sec:RL} but without any imitation pre-training. We adopt the \added{IFM}'s training strategy, including a terrain curriculum. However, we set the standard deviation to 3.0 and the PPO clip parameter to 0.2 and do not limit the exploration.
    \item \textbf{\added{Vanilla RL(CR):}} \added{The Vanilla RL(CR) baseline is trained similarly but with an additional contact biasing reward as in Sec.~\ref{sec:reward_formulation}.}
    \item \textbf{\added{Concurrent RL:}} \added{The Concurrent RL baseline is the exact work from Ji et al.~\cite{ji2022concurrent}.}
\end{itemize}
\subsubsection{Simulation}
We use RaiSim~\cite{raisim} as our physics simulator. The simulation time step is given as 0.001s, and the control frequency is set to $100$~Hz. We conduct the experiments on AMD Ryzen Threadripper 3970X 32-Core Processor CPU and GeForce RTX 3080 Ti GPU.
\subsubsection{Hardware}
We select the Mini Cheetah robot~\cite{katz2019mini} as the hardware platform, which is lightweight and capable of demonstrating dynamic locomotion.

\subsection{Simulation}
This section discusses the improved robustness and tracking performance of our method, particularly compared to the model-based expert controller.

\subsubsection{Robustness}

We evaluated the robustness of all the control policies on \replaced{three}{four} different terrains in simulation: rough, slippery\deleted{, stair}, and conveyor belts. Here, conveyor terrain is constructed such that several blocks relatively move in the $x$-direction. Please note that rough and slippery terrains are part of training environments, while conveyor belt terrain is unseen during training. We select the survival time until fall as a performance criterion\deleted{, except for the stair terrain in which we measure the normalized number of climbed stairs: $\{\text{\# of climbed stairs}\}/\{\text{\# of stairs}\}$}. 

\begin{figure}
\vspace{0.25cm}
\centering
\includegraphics[width=0.9\columnwidth]{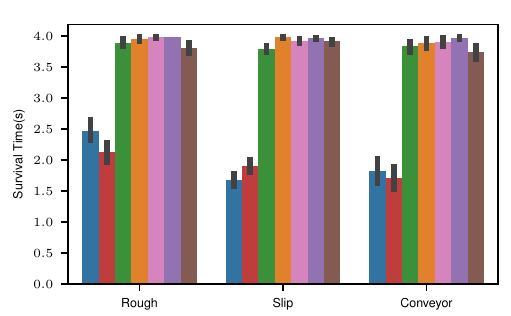}
\includegraphics[width=0.9\columnwidth]{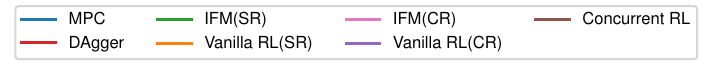}
\vspace{-1.0em}
\caption{
Comparison of the robustness in simulated environments.\deleted{ We measured either the survival time or the normalized number of climbed stairs.} \added{IFM} (ours) and \added{Vanilla RL} perform better than MPC and DAgger. \deleted{The grey-shaded area corresponds to the results on out-of-distribution of the parameters.}}
\label{fig:sim_robustness}
\vspace{-1.5em}
\end{figure}

The results are illustrated in Fig.~\ref{fig:sim_robustness}. Overall, \added{IFM} significantly improves the robustness of MPC and DAgger. This trend is not only for in-distribution tasks \added{(rough and slip terrains)} seen during training but also for a novel \added{conveyor} terrain.\deleted{with either the out-of-the-training distribution parameters (e.g., terrain factor $>$ 1) or new terrain types (conveyor belts).} These results imply that learning-based policies can effectively adapt the robot's gait\added{, particularly contact sequences,} to challenging terrains, while MPC's performance is limited due to its \replaced{predefined model}{ model assumptions}, such as a fixed contact sequence and reference swing leg trajectory. 

On the other hand, \added{IFM(SR)}'s performance is comparable to that of Vanilla RL\added{(SR)} and even slightly worse for some out-of-distribution tasks. This is because Vanilla RL\added{(SR)} found an asymmetric, energy-inefficient gait that is similar to pronking. \added{This discrepancy decreases as contact biasing reward is introduced and more extensive exploration is employed.} We will further compare the Vanilla RL and \added{IFM} policies in Sec.~\ref{sec:vsRL}.

\subsubsection{Tracking Performance}
We also compare the policies by measuring their capabilities to follow the velocity command on the flat terrain. The command vector consists of a 3-dimensional vector, $[v_x,v_y,\omega_z]$, which is sampled uniformly from $v_x \in [-0.3,1]$, $v_y \in [-0.5,0.5]$, and $\omega_z \in [-0.6,0.6]$. We changed this command four times in one trajectory.

\begin{table}[]
\caption{Simulation and Hardware Performance on the Flat Terrain}
\label{tab:overall_performance}
\renewcommand{\arraystretch}{2.}
\setlength{\tabcolsep}{2pt}
\resizebox{\columnwidth}{!}{%
\begin{tabular}{|c|cc|cccc|}
\hline
\multirow{2}{*}{Method} &
  \multicolumn{2}{c|}{Simulation} &
  \multicolumn{4}{c|}{Hardware Experiments} \\ \cline{2-7} 
 &
  \multicolumn{1}{c|}{\renewcommand{\arraystretch}{0.7}\begin{tabular}[c]{@{}c@{}}Track err (↓)\\ (linear)\end{tabular}} &
  \renewcommand{\arraystretch}{0.7}\begin{tabular}[c]{@{}c@{}}Track err (↓)\\ (angular)\end{tabular} &
  \multicolumn{1}{c|}{\renewcommand{\arraystretch}{0.7}\begin{tabular}[c]{@{}c@{}}Track err (↓)\\ (linear)\end{tabular}} &
  \multicolumn{1}{c|}{\renewcommand{\arraystretch}{0.7}\begin{tabular}[c]{@{}c@{}}Track err (↓)\\ (angular)\end{tabular}} &
  \multicolumn{1}{c|}{\renewcommand{\arraystretch}{0.7}\begin{tabular}[c]{@{}c@{}}CoT (↓)\\ (1m/s)\end{tabular}} &
  \renewcommand{\arraystretch}{0.7}\begin{tabular}[c]{@{}c@{}}PPI (↓)\\ (1m/s)\end{tabular} \\ \hline
MPC/WBIC &
  \multicolumn{1}{c|}{\renewcommand{\arraystretch}{0.7} \begin{tabular}[c]{@{}c@{}}0.1520 \\ (±0.0169)\end{tabular}\cellcolor[gray]{.8}} &
  \renewcommand{\arraystretch}{0.7} \begin{tabular}[c]{@{}c@{}}0.5335\\ (±0.0519)\end{tabular} &
  \multicolumn{1}{c|}{\renewcommand{\arraystretch}{0.7} \begin{tabular}[c]{@{}c@{}}\added{0.2334}\\ \added{(±0.0062)}\end{tabular}\cellcolor[gray]{.8}} &
  \multicolumn{1}{c|}{\renewcommand{\arraystretch}{0.7} \begin{tabular}[c]{@{}c@{}}\added{0.7901}\\ \added{(±0.0108)}\end{tabular}} &
  \multicolumn{1}{c|}{\renewcommand{\arraystretch}{0.7} \begin{tabular}[c]{@{}c@{}}\added{0.4374}\\ \added{(±0.0113)}\end{tabular}} &
  \textbf{\renewcommand{\arraystretch}{0.7} \begin{tabular}[c]{@{}c@{}}\added{1.715}\\ \added{(±0.3547)}\end{tabular}\cellcolor[gray]{.8}} \\ \hline
DAgger &
  \multicolumn{1}{c|}{\renewcommand{\arraystretch}{0.7} \begin{tabular}[c]{@{}c@{}}0.1838\\ (±0.0217)\end{tabular}} &
  \renewcommand{\arraystretch}{0.7} \begin{tabular}[c]{@{}c@{}}0.5214\\ (±0.0449)\end{tabular} &
  \multicolumn{1}{c|}{\renewcommand{\arraystretch}{0.7} \begin{tabular}[c]{@{}c@{}}\added{0.2694}\\ \added{(±0.0065)}\end{tabular}} &
  \multicolumn{1}{c|}{\renewcommand{\arraystretch}{0.7} \begin{tabular}[c]{@{}c@{}}\added{1.153}\\ \added{(±0.0075)}\end{tabular}} &
  \multicolumn{1}{c|}{\renewcommand{\arraystretch}{0.7} \begin{tabular}[c]{@{}c@{}}\added{0.4489}\\ \added{(±0.0020)}\end{tabular}} &
  \renewcommand{\arraystretch}{0.7} \begin{tabular}[c]{@{}c@{}}\added{1.880}\\ \added{(±0.3067)}\end{tabular} \cellcolor[gray]{.8}\\ \hline
\added{Vanilla RL(SR)} &
  \multicolumn{1}{c|}{\renewcommand{\arraystretch}{0.7} \begin{tabular}[c]{@{}c@{}}0.2017\\ (±0.0125)\end{tabular}} &
  \renewcommand{\arraystretch}{0.7} \begin{tabular}[c]{@{}c@{}}0.2757\\ (±0.0114)\end{tabular} &
  \multicolumn{1}{c|}{\renewcommand{\arraystretch}{0.7} \begin{tabular}[c]{@{}c@{}}\added{0.3814}\\ \added{(±0.0711)}\end{tabular}} &
  \multicolumn{1}{c|}{\renewcommand{\arraystretch}{0.7} \begin{tabular}[c]{@{}c@{}}\added{1.339}\\ \added{(±0.1195)}\end{tabular}} &
  \multicolumn{1}{c|}{\renewcommand{\arraystretch}{0.7} \begin{tabular}[c]{@{}c@{}}\added{1.287}\\ \added{(±0.0981)}\end{tabular}} &
  \renewcommand{\arraystretch}{0.7} \begin{tabular}[c]{@{}c@{}}\added{10.13}\\ \added{(±2.007)}\end{tabular} \\ \hline
\added{IFM(SR)} &
  \multicolumn{1}{c|}{\textbf{\renewcommand{\arraystretch}{0.7} \begin{tabular}[c]{@{}c@{}}0.1335\\ (±0.0163)\end{tabular}}\cellcolor[gray]{.8}} &
  \renewcommand{\arraystretch}{0.7} \begin{tabular}[c]{@{}c@{}}0.2509\\ (±0.0153)\end{tabular} \cellcolor[gray]{.8}&
  \multicolumn{1}{c|}{\textbf{\renewcommand{\arraystretch}{0.7} \begin{tabular}[c]{@{}c@{}}\added{0.1620}\\ \added{(±0.0103)}\end{tabular}}\cellcolor[gray]{.8}} &
  \multicolumn{1}{c|}{\renewcommand{\arraystretch}{0.7} \begin{tabular}[c]{@{}c@{}}\added{0.7320}\\ \added{(±0.0100)}\end{tabular}\cellcolor[gray]{.8}} &
  \multicolumn{1}{c|}{\renewcommand{\arraystretch}{0.7} \begin{tabular}[c]{@{}c@{}}\added{0.4237}\\ \added{(±0.0144)}\end{tabular}\cellcolor[gray]{.8}} &
  \renewcommand{\arraystretch}{0.7} \begin{tabular}[c]{@{}c@{}}\added{1.798}\\ \added{(±0.3975)}\end{tabular}\cellcolor[gray]{.8} \\ \hline
\added{Vanilla RL(CR)} &
  \multicolumn{1}{c|}{\renewcommand{\arraystretch}{0.7} \begin{tabular}[c]{@{}c@{}}\added{0.1766}\\ \added{(±0.0175)}\end{tabular}\cellcolor[gray]{.8}} &
  \renewcommand{\arraystretch}{0.7} \begin{tabular}[c]{@{}c@{}}\added{0.2926}\\ \added{(±0.0123)}\end{tabular} &
  \multicolumn{1}{c|}{\renewcommand{\arraystretch}{0.7} \begin{tabular}[c]{@{}c@{}}\added{0.2653}\\ \added{(±0.0057)}\end{tabular}} &
  \multicolumn{1}{c|}{\renewcommand{\arraystretch}{0.7} \begin{tabular}[c]{@{}c@{}}\added{0.9680}\\ \added{(±0.0146)}\end{tabular}} &
  \multicolumn{1}{c|}{\renewcommand{\arraystretch}{0.7} \begin{tabular}[c]{@{}c@{}}\added{0.9154}\\ \added{(±0.0317)}\end{tabular}} &
  \renewcommand{\arraystretch}{0.7} \begin{tabular}[c]{@{}c@{}}\added{5.001}\\ \added{(±0.6006)}\end{tabular} \\ \hline
\added{IFM(CR)}&
  \multicolumn{1}{c|}{\renewcommand{\arraystretch}{0.7} \begin{tabular}[c]{@{}c@{}}\added{0.1879}\\ \added{(±0.0175)}\end{tabular}} &
  \renewcommand{\arraystretch}{0.7} \begin{tabular}[c]{@{}c@{}}\added{0.2624}\\ \added{(±0.0102)}\end{tabular} \cellcolor[gray]{.8}&
  \multicolumn{1}{c|}{\renewcommand{\arraystretch}{0.7} \begin{tabular}[c]{@{}c@{}}\added{0.2826}\\ \added{(±0.0102)}\end{tabular}} &
  \multicolumn{1}{c|}{\renewcommand{\arraystretch}{0.7} \begin{tabular}[c]{@{}c@{}}\added{0.6978}\\ \added{(±0.0321)}\end{tabular}\cellcolor[gray]{.8}} &
  \multicolumn{1}{c|}{\renewcommand{\arraystretch}{0.7} \begin{tabular}[c]{@{}c@{}}\added{0.3250}\\ \added{(±0.0049)}\end{tabular}\cellcolor[gray]{.8}} &
  \renewcommand{\arraystretch}{0.7} \begin{tabular}[c]{@{}c@{}}\added{2.936}\\ \added{(±0.1979)}\end{tabular} \\ \hline
\added{Concurrent RL }&
  \multicolumn{1}{c|}{\renewcommand{\arraystretch}{0.7} \begin{tabular}[c]{@{}c@{}}\added{0.1793}\\ \added{(±0.0153)}\end{tabular}} &
  \textbf{\renewcommand{\arraystretch}{0.7} \begin{tabular}[c]{@{}c@{}}\added{0.2185}\\ \added{(±0.0089)}\end{tabular}\cellcolor[gray]{.8}} &
  \multicolumn{1}{c|}{\renewcommand{\arraystretch}{0.7} \begin{tabular}[c]{@{}c@{}}\added{0.1722}\\ \added{(±0.0155)}\end{tabular}\cellcolor[gray]{.8}} &
  \multicolumn{1}{c|}{\textbf{\renewcommand{\arraystretch}{0.7} \begin{tabular}[c]{@{}c@{}}\added{0.4099}\\ \added{(±0.0040)}\end{tabular}}\cellcolor[gray]{.8}} &
  \multicolumn{1}{c|}{\textbf{\renewcommand{\arraystretch}{0.7} \begin{tabular}[c]{@{}c@{}}\added{0.2206}\\ \added{(±0.0065)}\end{tabular}}\cellcolor[gray]{.8}} &
  \renewcommand{\arraystretch}{0.7} \begin{tabular}[c]{@{}c@{}}\added{2.385}\\ \added{(±0.1602)}\end{tabular} \\ \hline
\end{tabular}%
}
\end{table}

The results are summarized in Table~\ref{tab:overall_performance} \added{where we highlighted the top 3 performances with gray backgrounds}. \replaced{IFM(SR) and Concurrent RL show the best performance in linear and angular velocity tracking, respectively.}{Overall, CIRL shows the best tracking performance for both linear and angular velocity commands.} MPC and DAgger failed to follow the command and fell off 12\% of the tests, which never happened for both \added{IFM} and Vanilla RL. They also showed worse linear and angular velocity tracking performance than \added{IFM(SR)}. Interestingly, \replaced{Vanilla RL(SR)}{RL} demonstrated poor linear tracking performance, even worse than MPC or DAgger. This is mainly because it has excessive $z$-direction movements produced by its pronking motion. Despite being trained with the same reward formulation as Vanilla RL\added{(SR)}, \added{IFM(SR)} did not show that problem because of pre-training. \added{Vanilla RL(CR) showed better tracking than Vanilla RL(SR) by following specific contact sequences, but still worse than IFM(SR).}

\subsubsection{Computation Time} \added{IFM}  reduces MPC's computation time from 0.0154s to 0.0010s, thanks to its simple network architecture. This showcases the benefits of \added{IFM} in scalable learning and real-time control.

\begin{figure}
\centering
\includegraphics[width=\columnwidth,trim=13.5cm 0 0 0,clip]{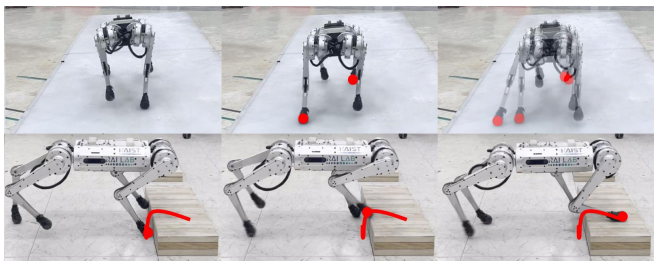}
\caption{ IFM policy can traverse slippery terrain with the friction of 0.22 (\textbf{Top}) and also overcome a 7.5cm step using foot-trapping reflexes (\textbf{Bottom}). }
\label{fig:slip_step_figure}
\vspace{-1em}
\end{figure}
\subsection{Hardware Experiments}
We further conducted hardware experiments\deleted{to compare the performance in the real world}.

\subsubsection{Robustness}
\begin{table}[]
\caption{Hardware Experimental Results on Challenging Terrains}
\centering
\renewcommand{\arraystretch}{1.5}
\resizebox{0.9\columnwidth}{!}{%
\begin{tabular}{|c|ccc|cc|}
\hline
\multirow{3}{*}{Method} & \multicolumn{3}{c|}{Step terrain}                           & \multicolumn{2}{c|}{\multirow{2}{*}{Slippery terrain(1m/s)}} \\ \cline{2-4}
       & \multicolumn{3}{c|}{Success rate(\%) $(\uparrow)$}                    & \multicolumn{2}{c|}{}              \\ \cline{2-6} 
                        & \multicolumn{1}{c|}{2.5cm} & \multicolumn{1}{c|}{5cm} & 7.5cm & \multicolumn{1}{c|}{Tracking error $(\downarrow)$} & Pass time(s) $(\downarrow)$\\ \hline
WBIC   & \multicolumn{1}{c|}{100} & \multicolumn{1}{c|}{100}   & 0  & \multicolumn{1}{c|}{Fail}   & Fail \\ \hline
DAgger & \multicolumn{1}{c|}{100} & \multicolumn{1}{c|}{0}   & 0  & \multicolumn{1}{c|}{Fail}   & Fail \\ \hline
\added{Vanilla RL(SR)}     & \multicolumn{1}{c|}{100} & \multicolumn{1}{c|}{100} & 60 & \multicolumn{1}{c|}{0.6550} & 2.63 \\ \hline
\added{IFM(SR)}   & \multicolumn{1}{c|}{100} & \multicolumn{1}{c|}{100} & 60 & \multicolumn{1}{c|}{\textbf{0.1893}} & \textbf{2.00} \\ \hline
\added{Vanilla RL(CR)}   & \multicolumn{1}{c|}{\added{100}} & \multicolumn{1}{c|}{\added{100}}   & \added{80}  & \multicolumn{1}{c|}{\added{Fail}}   & \added{Fail} \\ \hline
\added{IFM(CR)}   & \multicolumn{1}{c|}{\added{100}} & \multicolumn{1}{c|}{\added{100}}   & \added{\textbf{90}}  & \multicolumn{1}{c|}{\added{0.2396}}   & \added{2.550} \\ \hline
\added{Concurrent RL}   & \multicolumn{1}{c|}{\added{100}} & \multicolumn{1}{c|}{\added{0}}   & \added{0}  & \multicolumn{1}{c|}{\added{0.2081}}   & \added{2.335} \\ \hline
\end{tabular}%
}
\label{tab:hardware_improvement}
\vspace{-0.6cm}
\end{table}
First, we measured the success rate of stepping over the three individual steps whose heights were 2.5cm, 5cm, and 7.5cm, respectively (Table~\ref{tab:hardware_improvement}). \added{The success rate is measured over 10 trials. }WBIC could not step over the 7.5 cm height, which is beyond its foot clearance. The DAgger policy showed even worse performance and failed to overcome the 5 cm height. \added{IFM(CR) showed the best performance so far at a $90$~\% success rate (Fig.~\ref{fig:teaser} and Fig.~\ref{fig:slip_step_figure}).}

Further, we gave the robot a 1m/s command on the slippery terrain. We spread out the boric acid powder on a 6m whiteboard to create a terrain with a friction coefficient of 0.22 and measured the tracking error and the time taken to cross the central 2m area of the whiteboard. In the case of MPC and DAgger, Raibert's heuristic foothold could not maintain stability when the foot slipped significantly, resulting in a fall. Both \added{IFM(SR)} and \added{Vanilla RL(SR)} can pass the slippery surface, but \added{IFM(SR)} showed significant improvements over the tracking error and pass-time. \added{Similarly, IFM(CR) succeeded to pass this terrain in a reasonable time while Vanilla RL(CR) failed.}

\subsubsection{Tracking Performance}
\deleted{As in the simulation,} We measured the tracking performance on hardware, where each controller is tasked to follow the 6-second command (Fig.~\ref{fig:hardware_tracking})\added{ for five trials}. The \replaced{averaged errors}{results} are also summarized in Table~\ref{tab:overall_performance}.
\added{IFM(SR) and Concurrent RL are showing the lowest tracking errors in linear and angular velocity, respectively. This is notable in that WBIC is specifically designed for the hardware and Concurrent RL has complex rewards (nine terms) and an additional state estimation for better sim to real transferability. On the other hand, both Vanilla RL(SR) and Vanilla RL(CR) show worse tracking performance.}
\deleted{which indicates that the performance of CIRL is slightly better than that of WBIC. This is notable, considering that WBIC is specifically designed for flat terrain. On the other hand, RL showed much worse tracking performance.}

\subsection{Comparison between \added{IFM} and Vanilla RL} \label{sec:vsRL}
This section further highlights the strengths of \added{IFM} over Vanilla RL: (1) learning efficiency, (2) tracking performance, and (3) motion quality.
\subsubsection{Learning Efficiency}
\begin{figure}
\vspace{0.6cm}
\centering
\setlength{\tabcolsep}{0pt}
\resizebox{\columnwidth}{!}{%
\setlength{\arrayrulewidth}{1pt}
\begin{tabular}{ccc}
 \includegraphics[scale=1]{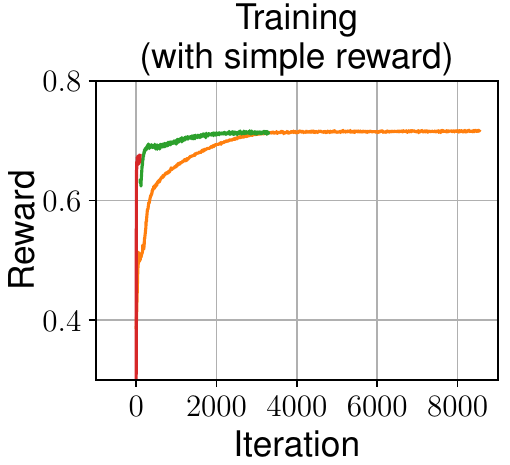} & \includegraphics[scale=1,trim={1.05cm 0 0 0},clip]{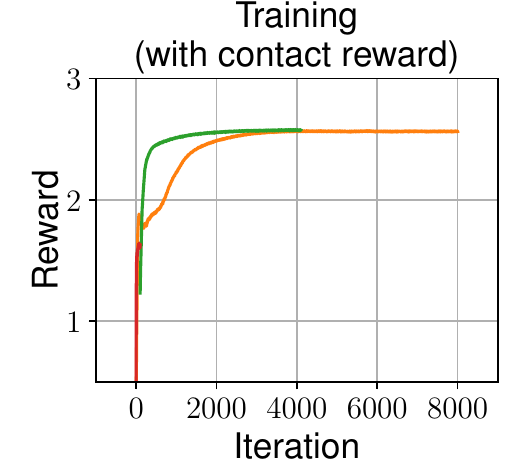}
\end{tabular}
}
\includegraphics[width=0.9\columnwidth]{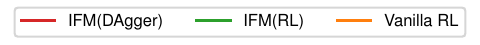}
\vspace{-0.4cm}
\caption{Learning curves of \added{Vanilla RL} and \added{IFM}, which illustrate faster convergence of \added{IFM} over Vanilla RL. \deleted{If we add one more challenging terrain, step stones, Vanilla RL performs worse than CIRL.}
}
\vspace{-0.4cm}
\label{fig:learning_curve}
\end{figure}

We compare Vanilla RL's and \added{IFM}'s learning efficiency by plotting the \deleted{normalized }reward in Fig.~\ref{fig:learning_curve}. The learning curves represent the mean of three trials, with the exclusion of confidence intervals due to their narrow range. Note that we divide the \added{IFM} learning curve into DAgger and RL stages. The plot clearly shows that \added{IFM} ($\sim2000$~iters) converges faster than Vanilla RL ($\sim4000$~iters). 

\deleted{In addition, we further evaluate the convergence on a harder problem, which includes one additional challenging terrain of step stones with fixed distances. In this case, Vanilla RL converges to the reward of 0.65, which is lower than the reward of CIRL, 0.7, even after much longer iterations. 
In our observation, Vanilla RL tends to find a pronking-like gait without extensively hand-tuning the reward function, which is unsuitable for step stones. }

\subsubsection{Tracking Performance}
\begin{figure}
\vspace{0.3cm}
\centering
\setlength{\tabcolsep}{0pt}
\renewcommand{\arraystretch}{0.}
\resizebox{\columnwidth}{!}{%
\begin{tabular}{ccc}
\includegraphics[scale=0.3]{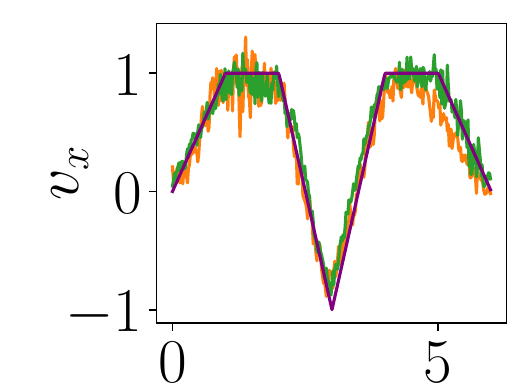}       & \includegraphics[scale=0.3,trim={0.6cm 0 0 0},clip]{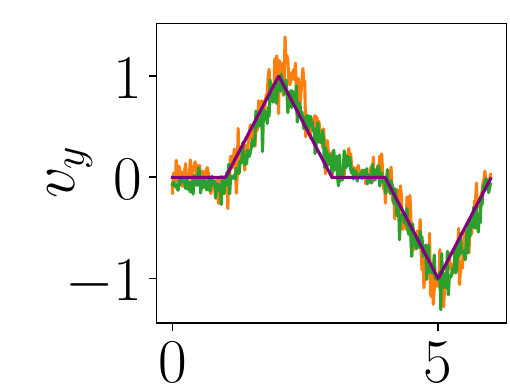}       & \includegraphics[scale=0.3]{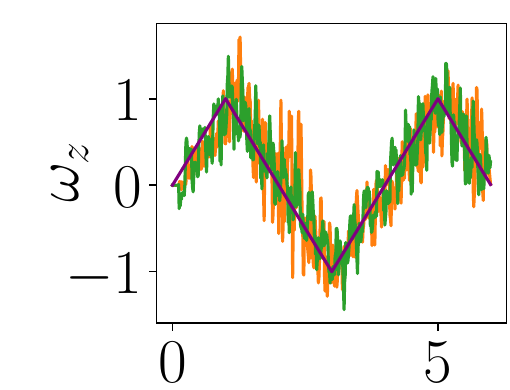}       \\
\includegraphics[scale=0.3]{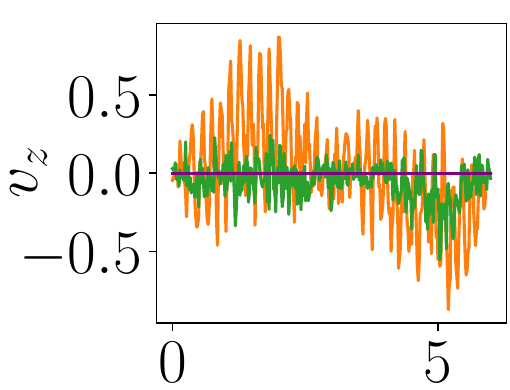}       & \includegraphics[scale=0.3,trim={0.6cm 0 0 0},clip]{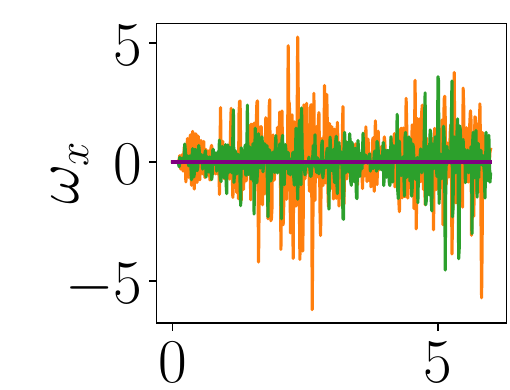}       & \includegraphics[scale=0.3]{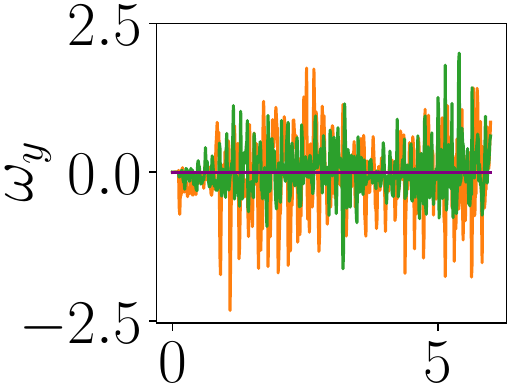}       \\
\multicolumn{3}{c}{\small Time(s)}
\end{tabular}%
}
\includegraphics[]{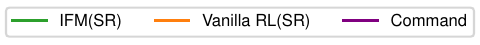}
\vspace{-0.3cm}
\caption{
Hardware command tracking performance. Overall, \added{IFM(SR)} shows better tracking than Vanilla RL\added{(SR)}, particularly in $v_z$, $\omega_x$, and $\omega_y$.}
\label{fig:hardware_tracking}
\vspace{-0.5cm}
\end{figure}

In our experiments, Vanilla RL\deleted{'s pronking-like gait} tends to show great robustness, as shown in \deleted{Tab.~\ref{tab:overall_performance}}\added{Fig.~\ref{fig:sim_robustness}} 
and Table~\ref{tab:hardware_improvement}. However, it is not well suited for velocity tracking and shows consistently lower tracking performances in both simulation and real-world environments. We plot the illustrative scenario \added{for simple reward policies }in Fig.~\ref{fig:hardware_tracking}, which shows worse tracking of Vanilla RL(SR) in $v_z$,  $\omega_x$, and $\omega_y$ directions due to its unnecessary movements. It will be possible to correct this behavior by comprehensive manual reward shaping\added{, such as Vanilla RL(CR)}: however, \added{IFM} offers a more structured approach to obtain better gaits.

\subsubsection{Motion Quality}

\begin{figure}
\vspace{0.25cm}
\centering
\includegraphics[width=0.9\columnwidth, scale=0.5]{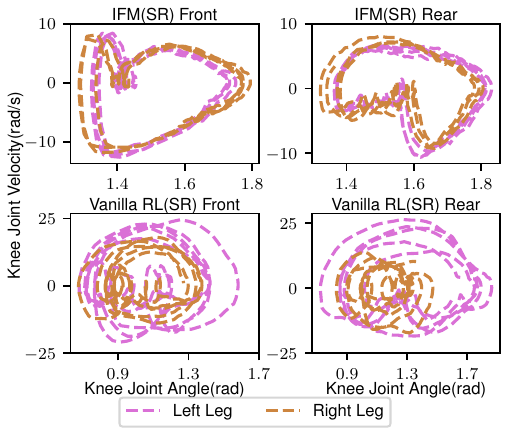}
\vspace{-0.4cm}
\caption{
 Hardware knee joint phase-portrait. \added{IFM(SR)}'s left leg and right leg have mostly overlapping phase-portrait, meaning that it has a periodic and symmetric motion. On the other hand, \added{Vanilla RL(SR)} shows highly asymmetric motions with rarely overlapping phase-portrait.}
\label{fig:hardware_PPI}
\vspace{-1em}
\end{figure}
We conduct a more in-depth analysis of the motion quality of \added{IFM} and Vanilla RL by measuring the Cost of Transportation (CoT) and Phase-Portrait Index (PPI)~\cite{hsiao2010review} on hardware. As shown in Table~\ref{tab:overall_performance}, \added{IFM methods'} CoT and PPI are comparable to those of MPC/WBIC. This is not surprising because \added{IFM} can be viewed as the finetuned version of a model-based expert. On the other hand, Vanilla RL policies exhibit very poor CoT and PPI, significantly larger than the others. The phase portrait in Fig.~\ref{fig:hardware_PPI} shows clear differences in PPI\added{ in simple reward setting}. In addition, the left and right phase portraits of \added{IFM(SR)} are nearly overlapped, which indicates near-perfect symmetricity.

Our experimental results do not indicate that vanilla RL cannot discover symmetric and efficient gaits. However, it often requires task-specific reward shaping, as a successful reward function in one scenario can exhibit hugely different motions in other situations. \added{IFM} provides a more explicit method for regulating the style of motion and significantly alleviates the challenge of reward engineering.

\subsection{\added{Qualitative comparisons of MPC+RL methods}}
\begin{table}[]
\centering
\caption{Qualitative Comparison of MPC+RL methods}
\label{tab:qualitative}
\renewcommand{\arraystretch}{2.3}
\setlength{\tabcolsep}{2pt}
\resizebox{\columnwidth}{!}{%
\begin{tabular}{|c|c|c|c|c|}
\hline

 &
  \added{IFM (ours)} & \renewcommand{\arraystretch}{0.9}\begin{tabular}[c]{@{}c@{}}\added{Contact}\\ \added{Adaptive~\cite{da2021learning}}\end{tabular} & \added{GLiDE~\cite{xie2023glide}} & \added{MPC-net~\cite{carius2020mpc}} \\ \hline
\renewcommand{\arraystretch}{0.9}\begin{tabular}[c]{@{}c@{}} \added{MPC+RL} \\ \added{approach}\end{tabular} &
  \added{Imitation} &
  \added{Hierarchical} &
  \added{Hierarchical} &
  \added{Imitation} \\ \hline
\renewcommand{\arraystretch}{0.9}\renewcommand{\arraystretch}{0.9}\begin{tabular}[c]{@{}c@{}} \added{Network} \\ \added{Action Space}\end{tabular} &
  \added{Target joint angle} &
  \added{Contact sequences} &
  \added{Body acceleration} &
  \renewcommand{\arraystretch}{0.9}\begin{tabular}[c]{@{}c@{}}\added{Joint velocities} \\ \added{and foot contact} \\ \added{forces}\end{tabular} \\ \hline
\renewcommand{\arraystretch}{0.9}\begin{tabular}[c]{@{}c@{}}\added{Computation}\\ \added{Time}\end{tabular} &
  \textbf{\added{Low} (\textless{}1ms)} &
  \added{High} &
  \added{High} &
  \textbf{\added{Low} (\textless{}1ms)} \\ \hline
\renewcommand{\arraystretch}{0.9}\begin{tabular}[c]{@{}c@{}}\added{Contact} \\ \added{Adaptation}\end{tabular} &
  \textbf{\added{Yes} }&
  \textbf{\added{Yes} }&
  \added{No} &
  \added{Yes} \\ \hline
\renewcommand{\arraystretch}{0.9}\begin{tabular}[c]{@{}c@{}}\added{Improvement of} \\\added{ MPC}\end{tabular} &
  \textbf{\added{Yes} }&
  \textbf{\added{Yes} }&
  \textbf{\added{Yes} }&
  \added{No} \\ \hline
\added{Demo terrain}\footnotemark &
  \added{F,R,Sl,C,St }&
  \added{F,Sl,C }&
  \added{F} &
  \added{F} \\ \hline
\end{tabular}%
}
\vspace{-0.4cm}
\end{table}
\added{To provide additional contextualization of our work, we present qualitative comparison in Table~\ref{tab:qualitative}. In comparison to other studies, our proposed method exhibits faster computation time and more robust foot adaptation. This advantage stems from the absence of online optimization and the lack of predefined swing leg controls in IFM. Furthermore, we have demonstrated the improved performance of IFM through finetuning its expert MPC, as evidenced by our extensive evaluation in simulation and on hardware.}
\footnotetext{Demonstrated terrains in each paper. F: flat terrain, R: rough terrain, Sl: slippery terrain (or banana peel), C: conveyor belt (or treadmill), St: step terrain.}
\section{CONCLUSIONS}
We propose a novel \added{Imitating and Finetuning Model Predictive Control} framework to obtain an effective locomotion controller with periodic and symmetric gaits. We first develop a model-based controller using DDP and the Raibert heuristic. Then we train an imitation policy using behavior cloning and further finetune it on challenging terrains via deep RL. 
By finetuning with RL, \added{IFM} can have significantly improved robustness compared to its expert policy (MPC) on challenging terrains and also reduces the computation time. By pre-training with MPC, \added{IFM} can find better gaits with lower CoT and PPI in a sample-efficient manner compared to Vanilla RL. We conduct comprehensive experiments in both simulation and hardware environments to highlight the benefits of the proposed method.

One future research direction will be to extend the proposed method for vision-based locomotion so that the policy can actively negotiate the obstacles. We expect that our approach can greatly contribute to more sample-efficient and effective learning for visual locomotion as well. However, we may need to carefully design the imitation learning stage to support more complicated network architectures, such as Long short-term memory or convolutional neural network.


\section*{ACKNOWLEDGMENT}

{This work was supported in part by Korea Evaluation Institute of Industrial Technology (KEIT) funded by the Korea Government (MOTIE) under Grant No.20018216, Development of mobile intelligence SW for autonomous navigation of legged robots in dynamic and atypical environments for real application.
Also, this work was supported by MIT Biomimetic Robotics Lab, NAVER LABS.}
\bibliographystyle{bibtex/IEEEtran}
\bibliography{bibtex/IEEEabrv,bibtex/ref}

\addtolength{\textheight}{-12cm}   


\end{document}